\documentclass[runningheads]{llncs}

\usepackage{eccv}

\usepackage{eccvabbrv}

\usepackage{graphicx}
\usepackage{booktabs}

\usepackage[accsupp]{axessibility}  

\usepackage{hyperref}

\usepackage{orcidlink}

\begin{document}

\title{Enhancing Industrial Transfer Learning with Style Filter: Cost Reduction and Defect-Focus}

\titlerunning{Style Filter}

\author{Chen Li\inst{1,2}\orcidlink{0000-0002-6749-7416} \and
Ruijie Ma\inst{1} \and
Xiang Qian\inst{1} \and
Xiaohao Wang\inst{1} \and
Xinghui Li\inst{1,2}}

\authorrunning{Li et al.}

\institute{
Shenzhen International Graduate School, Tsinghua University, China \and
Tsinghua-Berkeley Shenzhen Institute, Tsinghua University, China}

\maketitle

\begin{abstract}
  Addressing the challenge of data scarcity in industrial domains, transfer learning emerges as a pivotal paradigm. This work introduces Style Filter, a tailored methodology for industrial contexts. By selectively filtering source domain data before knowledge transfer, Style Filter reduces the quantity of data while maintaining or even enhancing the performance of transfer learning strategy. Offering label-free operation, minimal reliance on prior knowledge, independence from specific models, and re-utilization, Style Filter is evaluated on authentic industrial datasets, highlighting its effectiveness when employed before conventional transfer strategies in the deep learning domain. The results underscore the effectiveness of Style Filter in real-world industrial applications.
  \keywords{Industrial application \and Transfer learning \and Style transfer \and Defect detection}
\end{abstract}

\section{Introduction}
\label{sec:intro}

Surface defect detection holds pivotal significance in industrial contexts, ensuring product quality, safety, and reliability \cite{xie2016novel, yang2016defect}. Recent advancements in deep learning algorithms have brought about a paradigm shift in computer vision, particularly in image analysis tasks. These algorithms exhibit exceptional proficiency in extracting high-level features and discerning intricate patterns from complex visual data. Integration of deep learning with surface defect detection caters to the imperative for robust and precise defect identification in an automated and efficient manner \cite{nath2023nslnet, singh2023automated, li2023deep, usamentiaga2022automated, li2023dayolov5}.

A critical challenge lies in acquiring sufficiently large datasets to support the training of deep neural networks. Singh et al. (2023) explored the feasibility of utilizing publicly available labeled datasets and pre-trained Convolutional Neural Networks (CNNs) for surface defect detection algorithms. However, they found that such datasets are often insufficient for specific industrial detection tasks, such as those involving machined surfaces of flat washers and tapered rollers \cite{singh2023automated}. Consequently, enhancing model performance with limited samples has become a primary research objective.

In industrial settings where data scarcity is prevalent, it is a great choice to consider leveraging existing datasets from other manufacturers or previous batches. Typically, transfer learning is employed to utilize the source domain dataset to serve the task of the target domain. However, a large amount of unfiltered source domain data may potentially have a negative impact on the model performance. Therefore, in this work, we propose a method called Style Filter, which selectively filters source domain data before knowledge transfer. This approach reduces the cost of utilizing source domain data while ensuring that the effectiveness of the transfer strategy does not decrease and may even improve.

\section{Related Work}

\subsection{Industrial Surface Defect Detection}

In recent years, the fusion of machine vision with deep learning algorithms has gained considerable traction for industrial surface defect detection applications. This trend is driven by the benefits of heightened prediction accuracy, accelerated inspection speeds, and cost-effective integration into production lines \cite{jha2023deep, singh2023comparative}.

In the domain of industrial surface defect detection, research efforts predominantly focus on specific and well-defined structures within particular tasks \cite{tian2022dcc, xie2020ffcnn, xu2021automatic, dong2019pga, uzen2023multi}. For instance, Dong et al. (2019) proposed a pyramid feature fusion module and a global context attention module to mitigate issues stemming from significant appearance differences among defects across classes \cite{dong2019pga}. Similarly, Uzen et al. (2023) employed depthwise separable convolution layers to reduce parameters and introduced the 3D spectral and spatial features extraction (3DFE) module for extracting comprehensive spectral, spatial, and semantic features. Additionally, they incorporated the multi-input attention gate (MIAG) module to prioritize critical details \cite{uzen2023multi}.

However, a critical challenge lies in acquiring sufficiently large datasets to support the training of deep neural networks. To address the limitations imposed by small sample sizes, researchers often turn to generative techniques for data augmentation \cite{ma2023shape, wei2020simulation, wei2020defective, arikan2019surface, situ2021automated}. These methods generate synthetic data to supplement sparse training samples, thereby enhancing the model's learning capacity and generalization performance under data-limited scenarios. Few-shot learning also emerges as a prevalent strategy in such contexts. Shi et al. (2023) achieved classification by comparing positive and negative samples, demonstrating robust generalization capabilities to new samples \cite{shi2023few}. Moreover, domain adaptation has emerged as a viable approach. Li et al. (2023) integrated a domain adaptive module into the YOLOv5 network, resulting in a trained model exhibiting promising performance across source and target domains \cite{li2023dayolov5}.

\subsection{Style Transfer}

Style transfer, a technique to change an image's visual style while keeping its content, gained attention with Gatys et al.'s (2016) neural algorithm for artistic style \cite{gatys2016image}. This method separates content and style in natural images. Li et al. (2017) \cite{li2017demystifying} explained this approach with math, showing that aligning feature map Gram matrices is like minimizing Maximum Mean Discrepancy (MMD) with a specific kernel. They argue that the key to neural style transfer is matching feature distributions between style and generated images.

Huang et al. (2017) introduced Adaptive Instance Normalization (AdaIN) \cite{huang2017arbitrary}, adapting the channel-wise mean and variance of the feature maps to style features, enhancing flexibility. Nam et al. (2018) and Geirhos et al. (2018) \cite{nam2018batch, geirhos2018imagenet} highlighted style's importance in tasks. Li et al.'s work \cite{li2017demystifying} revealed the link between domain adaptation and style transfer, leading to various applications. Atapour et al. (2018) \cite{atapour2018real} used style transfer and adversarial training for monocular depth estimation from real-world images. They trained on synthetic data, adapting via style transfer for real-world scenarios. 

In defect detection, style transfer aids data generation. Some studies create new datasets by changing image styles \cite{rodriguezdomain, ma2023shape, wang2022robust}, while others add defects \cite{wei2020simulation} or diverse backgrounds \cite{wei2020defective}. The augmentation of styled datasets enhances network robustness; however, the increase in data also implies an escalation in training time costs.

\section{Methodology}
\label{sec:method}

\subsection{Basic Idea}
Before delving into the introduction of our Style Filter method, it is essential to first outline its basic idea. Inspired by their work \cite{li2017demystifying}, we discovered that style can effectively characterize the distinctions among samples from different sources. Experiments are conducted to validate this notion and visualized the results accordingly.

\begin{figure}[tb]
  \centering
  \includegraphics[width=.8\textwidth]{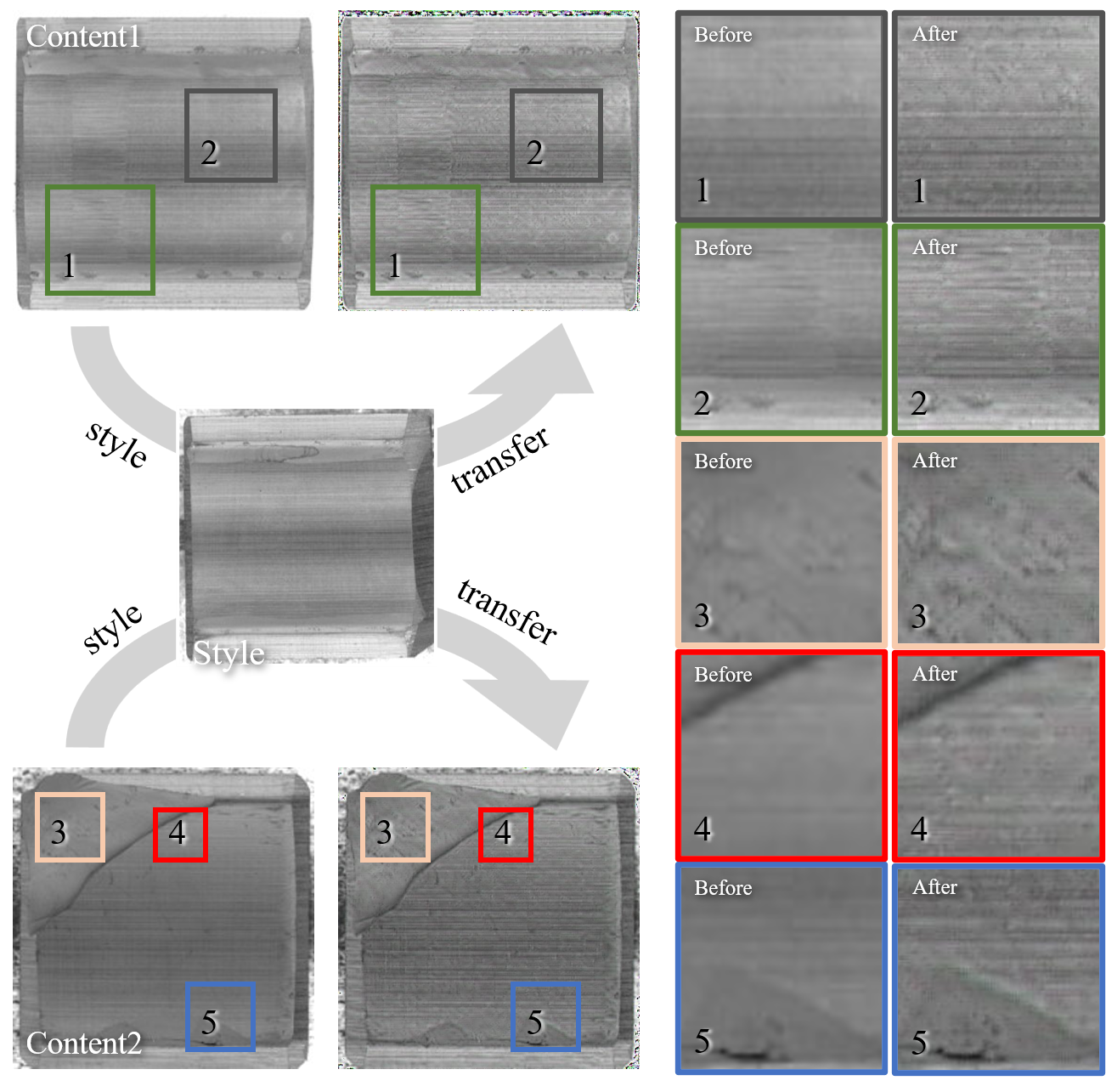}
  \caption{Basic idea of our Style Filter (SF): style can be employed to describe the differences between samples from various sources. The idea is verified by the visual outcomes of a style transfer experiment conducted on magnetic tile samples collected from various sources. (Best viewed in color.)}
  \label{fig:style}
\end{figure}

\cref{fig:style} presents the visual outcomes of a style transfer experiment conducted on magnetic tile datasets collected from various sources. This visualization underscores the discernible differences between samples originating from different sources, which can be delineated through stylistic variations. The figure showcases the impact of style transfer on magnetic tile samples, revealing variations in style. Specifically, it juxtaposes two magnetic tile sample images, illustrating the comparison before and after style transfer using a different style image. Highlighted colored boxes focus on local zoomed-in regions, depicting changes in the content images pre- and post-style transfer. The gray (\#1), yellow (\#3), and predominant areas within the red box (\#4) signify successful learning of texture, complexity, and other stylistic features. Conversely, the green box (\#2), the top-right corner of the red box (\#4), and the bottom-left corner of the blue box (\#5) emphasize preserved or even accentuated defect features.

As illustrated in \cref{fig:style}, the style transfer process enables the assimilation of texture and complexity from the style image into the content image while retaining or enhancing defect features. Additionally, when the style image exhibits pronounced contrast levels, the content sample inherits these specific contrast characteristics, thereby augmenting the contrast of the content image post-style transfer. Consequently, enhanced contrast facilitates improved detection of defect areas and fosters more intuitive human judgment of defects.

\subsection{Style Filter}

Since the variations between products from different sources can be described using style, style could be employed to filter the source domain dataset. 

\begin{figure}[tb]
  \centering
  \includegraphics[width=.98\textwidth]{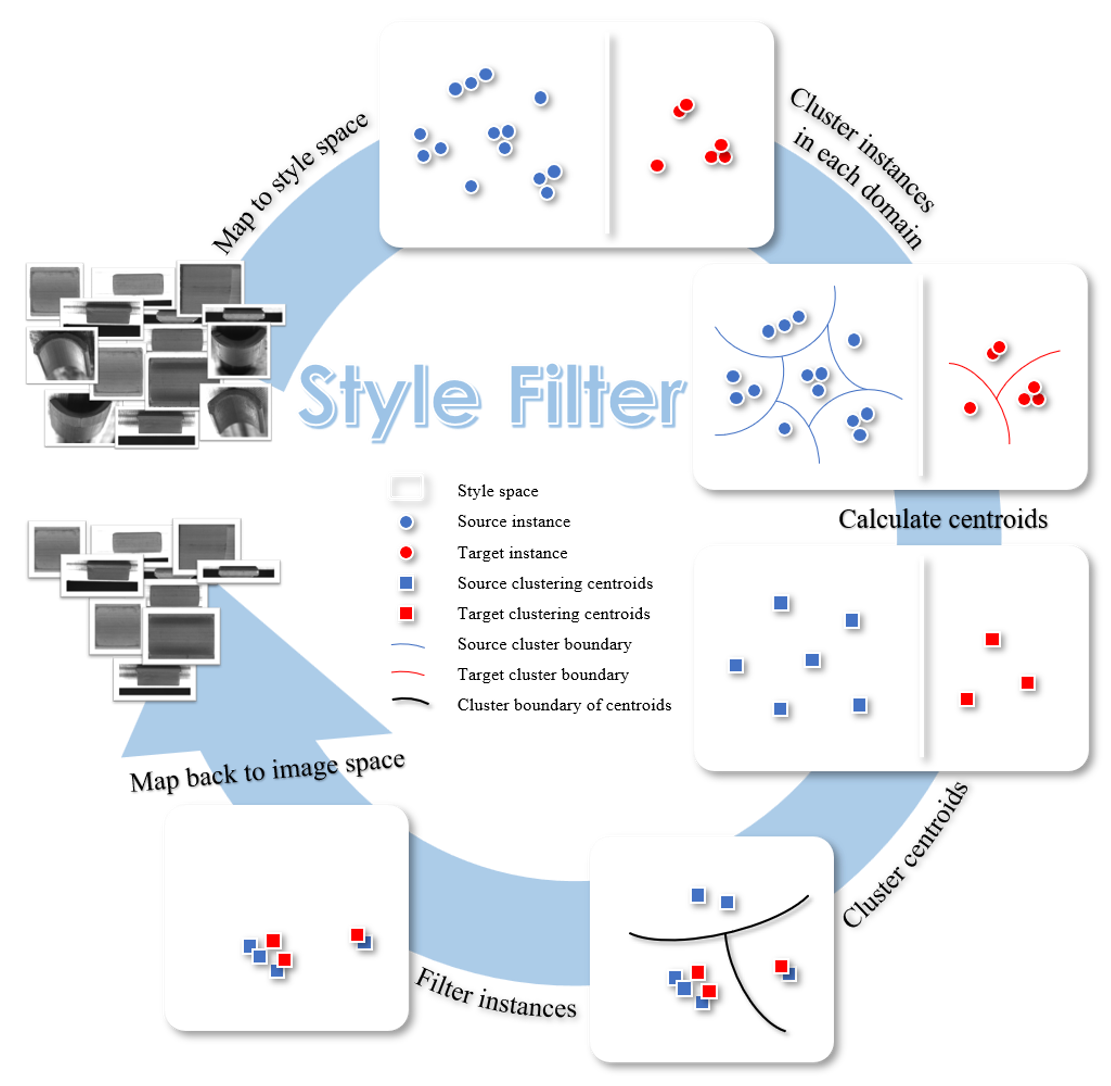}
  \caption{Overall flowchart of our Style Filter.}
  \label{fig:sf}
\end{figure}

The Style Filter (SF), specifically designed for devices and defects of identical types originating from diverse sources within the industrial domain, is structured into six steps, as illustrated in \cref{fig:sf}: 1) mapping to style spaces, 2) clustering instances within each domain, 3) calculating centroids of each cluster, 4) clustering centroids, 5) filtering instances, and 6) mapping back to image spaces. 

\begin{figure}[tb]
  \centering
  \includegraphics[width=.55\textwidth]{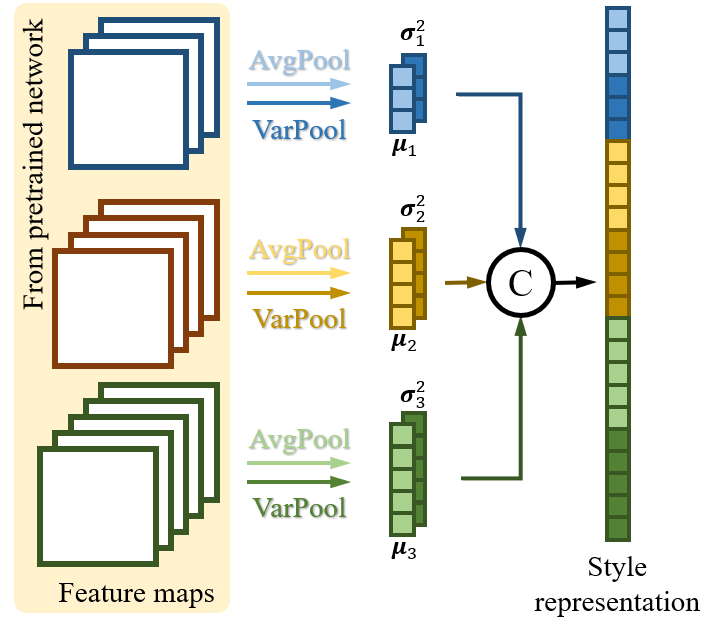}
  \caption{Specific structure of mapping images to style space. The structure can be primarily divided into two steps: extracting style features and feature fusion.}
  \label{fig:image2style}
\end{figure}

Mapping images to the style space is a crucial step in this process, as detailed in \cref{fig:image2style}. Initially, a pre-trained model on a public dataset is used to process both the source and target domain data. This results in obtaining different depth feature maps for each image. Subsequently, inspired by \cite{huang2017arbitrary}, the channel-wise mean and variance are computed for each feature map as style vectors, and all the style vectors are concatenated into a one-dimensional tensor. This tensor serves as the style representation for the input image. Consequently, each image is associated with a unique style representation. Moreover, employing the same pre-trained network and style extraction structure ensures that all images are mapped to the same style space.

In the work of Li et al. (2023) \cite{li2023dayolov5}, different depth feature maps are initially adjusted in channel numbers through a convolutional layer with a kernel size of one. This adjustment is solely aimed at ensuring that the features extracted at different depths have the same size when summed up later. However, as shown in Figure 3, our approach to extracting style features differs from that of DAYOLOv5 \cite{li2023dayolov5}. In our method, we do not adjust the channel numbers of feature maps at different depths, allowing the style information to be better preserved for subsequent clustering tasks. Additionally, concatenating all the extracted channel-wise mean and variance values into a one-dimensional tensor serves the same purpose.


After all images have been mapped to the same style space, clustering algorithms are employed to cluster samples from both the source and target domains, and their cluster centroids are computed. Once the cluster centroids for both domains are obtained, all centroids are clustered again using the same clustering algorithm. Subsequently, instances corresponding to source domain cluster centroids that are not classified into the same cluster as the target domain cluster centroids will be removed. Finally, leveraging the one-to-one mapping between sample images and the instances in the style space, the source domain data that will be taken into future transfer learning training is obtained.

In SF, style representations mapping from the source and target domains are clustered separately in the style space, enabling the batch processing of large amounts of data and the selection of whether to retain or remove them. Although this strategy is coarse, its deployment is extremely straightforward. Moreover, statistical stability is sought through batch processing, making it less susceptible to outliers. 
Moreover, during the clustering process, a larger number of clusters, denoted by $k$, is recommended. This allows for a more detailed partitioning and subsequent filtering, enhancing the granularity of the process. However, increasing the number of clusters incurs additional cost, necessitating the identification of the optimal $k$ value that truly suits the specific application scenario.

SF operates independently of subsequent tasks, making it naturally less dependent on specific network architectures and applicable to a wider range of applications. Additionally, the results of each step of SF can be stored, making it highly reusable for sequential tasks. Furthermore, apart from the need for additional selection of the cluster number $k$, all other steps require no prior knowledge or sample labels, greatly reducing the costs of training and annotation and minimizing the need for prior knowledge.

\section{Experiments}

\subsection{Datasets}

\begin{figure}[tb]
  \centering
  \begin{subfigure}{0.38\linewidth}
    \includegraphics[width=\textwidth]{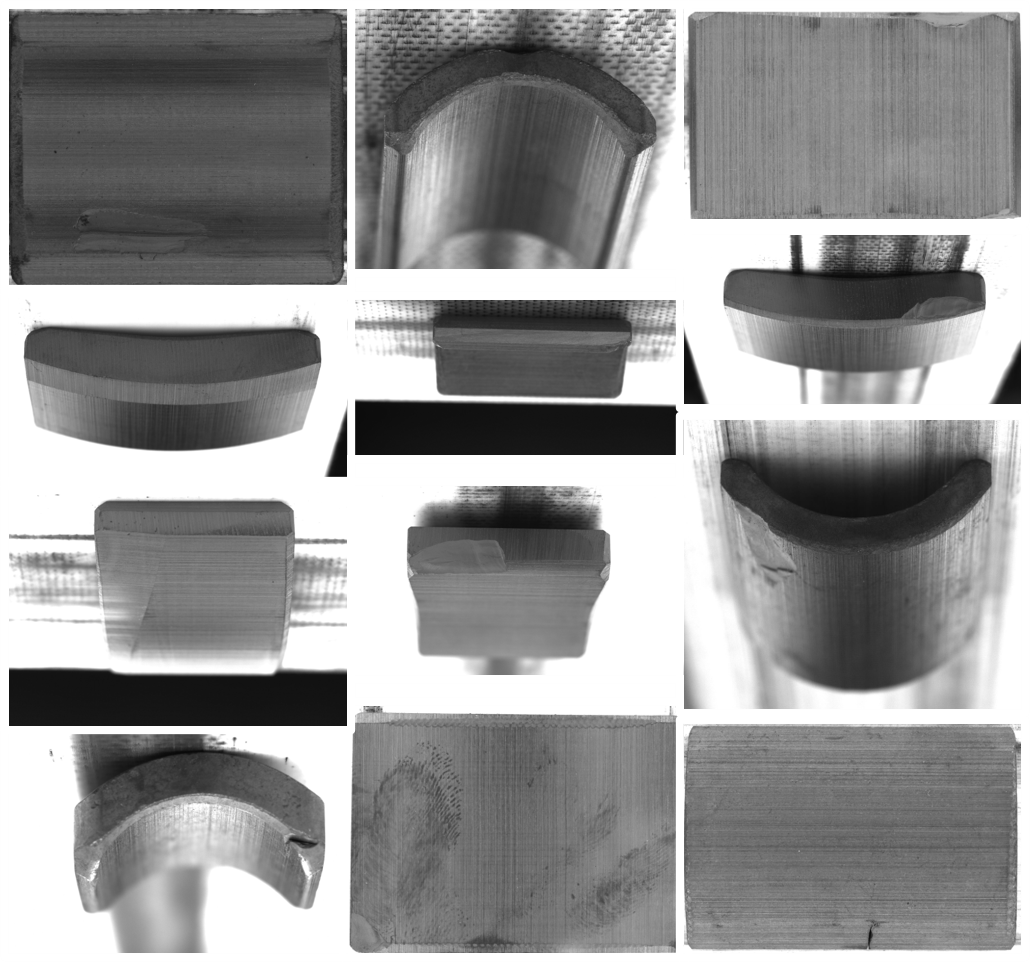}
    \caption{Different manufacturer samples.}
    \label{fig:diff_manu}
  \end{subfigure}
  \hfill
  \begin{subfigure}{0.56\linewidth}
    \includegraphics[width=\textwidth]{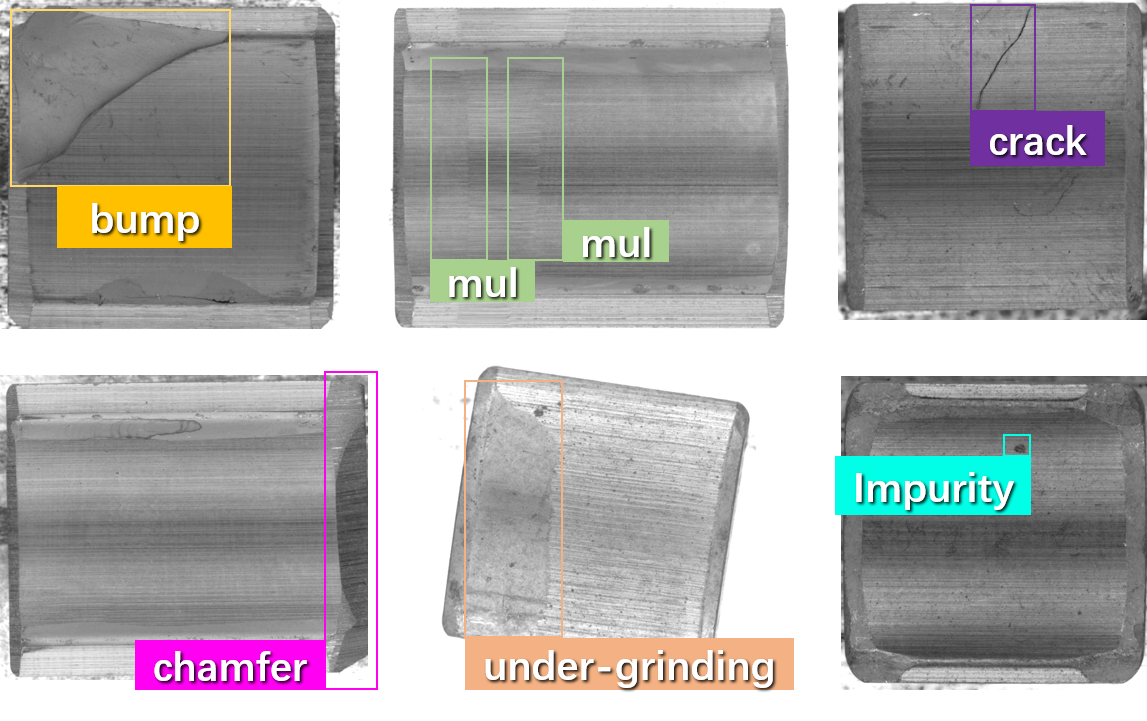}
    \caption{Surface defect types.}
    \label{fig:defects}
  \end{subfigure}
  \caption{Samples from different manufacturers and the surface defects. These samples manifest numerous discrepancies across several dimensions: actual shape variations of magnetic tiles, variations in backgrounds, and differences in contrast stemming from diverse shooting angles employed to capture distinct magnetic tile surfaces.}
  \label{fig:datasets}
\end{figure}

In this investigation, we opt for a \textbf{self-constructed magnetic tile dataset} as our experimental dataset, compiled from four distinct factory production lines. Samples sourced from different manufacturers showcase disparities across multiple dimensions, including photography environment (e.g., lighting, contrast), shooting angles (e.g., background), and sample arrangement (e.g., texture direction). \cref{fig:diff_manu} delineates these distinctions among samples originating from diverse sources.

During the dataset construction, only the most basic cleaning strategy was employed, primarily focusing on class filtering, to better replicate the situations encountered in real industrial production. Through this strategy, the aim was to capture and reproduce the challenges faced in actual production scenarios. Significant advantages are exhibited by our dataset over currently available publicly accessible datasets in terms of quantity, quality, and diversity to real-world scenarios in the industrial domain. By conducting experiments and evaluations using our dataset, the performance of our methods in the task of industrial defect detection can be comprehensively validated. Additionally, the inclusion of data that is not entirely "perfect" adds to the challenge of our task.

Six categories of surface defects are contained within our dataset: crack, bump, multifaceted, chamfer, impurity, and grind. The specific characteristics of these six defects are illustrated in \cref{fig:defects}.

\subsection{Implementation Details}

The magnetic tile datasets used in our study originate from four different factories, with one serving as the target domain dataset and the remaining three as the source domain datasets. \cref{table:cat_distr_data} presents the distributions of defect types in both the source and target domains. Our objective is to train a detector using the source domain datasets, supplemented with a small amount of labeled data from the target domain, and evaluate the defect detection performance on the target domain's test set before and after implementing SF. The labeled target domain dataset was obtained through random sampling to simulate real-world industrial scenarios. The size of the labeled target data will be abbreviated as LTS and indicated in subsequent tables.

\begin{table}[tb]
  \caption{Categorical distribution of the source and target domains.}
  \label{table:cat_distr_data}
  \centering
  \begin{tabular}{@{}l l l l l l l l@{}}
    \toprule
    Defects & crack & bump & multifaceted & chamfer & impurity & grind & total\\
    \midrule
    Source & 5627 & 14917 & 4205 & 708 & 277 & 863 & 21700\\
    Target & 1104 & 2432 & 1526 & 715 & 453 & 691 & 5927\\
  \bottomrule
  \end{tabular}
\end{table}

In the feature extraction part of SF, VGG-19 is chosen as our pre-trained feature extraction network. This choice was motivated not only by the model's early adoption in style transfer \cite{gatys2016image} but also by its low parameter count and fast inference speed, which align well with the efficiency requirements of industrial scenarios. Additionally, for the clustering algorithm, we utilized the K-means algorithm, which is widely used in deep learning, including in YOLO series detection networks \cite{redmon2017yolo9000}. Its simplicity, efficiency, and ability to rapidly process large amounts of data are highly desirable for industrial settings. However, selecting the number of clusters, the value of $k$, requires some prior knowledge. In our experiments, Elbow and Silhouette methods will be used to determine the optimal $k$ value selection. Also, PCA and t-SNE will be employed to visualize the clustering results \cite{mackiewicz1993principal, van2008visualizing}. 

Considering the pressing need for high-speed inference in industrial settings, single-stage detectors are the preferred choice \cite{cai2021yolov4, zhou2023ssda}. Hence, in our study validating the efficacy of Style Filter (SF) in surface defect detection tasks, the YOLOv7-series models are employed. The training procedure follows the settings outlined in \cite{wang2023yolov7}, ensuring consistency across the YOLOv7-series detectors. Furthermore, we adopt an early stopping strategy to maximize the utilization of the YOLOv7-series models' capabilities. 

During the testing phase, detector performance is assessed by reporting the mean average precision (mAP) at an Intersection over Union (IoU) threshold of 0.5 (referred to as mAP@0.5) and the mean mAP value across IoU thresholds ranging from 0.5 to 0.95 (referred to as mAP@0.5:0.95). To ensure the robustness and generalizability of the results, all experiments are repeated multiple times and averaged.

\subsection{Style Filter Results}

\begin{figure}[tb]
  \centering
  \begin{subfigure}{0.48\linewidth}
    \includegraphics[width=\textwidth]{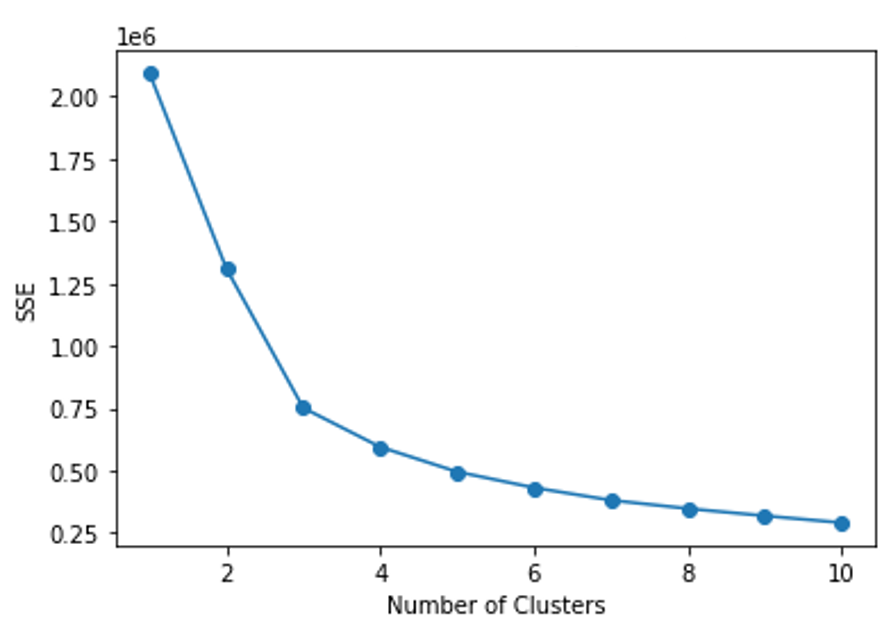}
    \caption{Elbow method}
    \label{fig:cluster_src_sse}
  \end{subfigure}
  \hfill
  \begin{subfigure}{0.48\linewidth}
    \includegraphics[width=\textwidth]{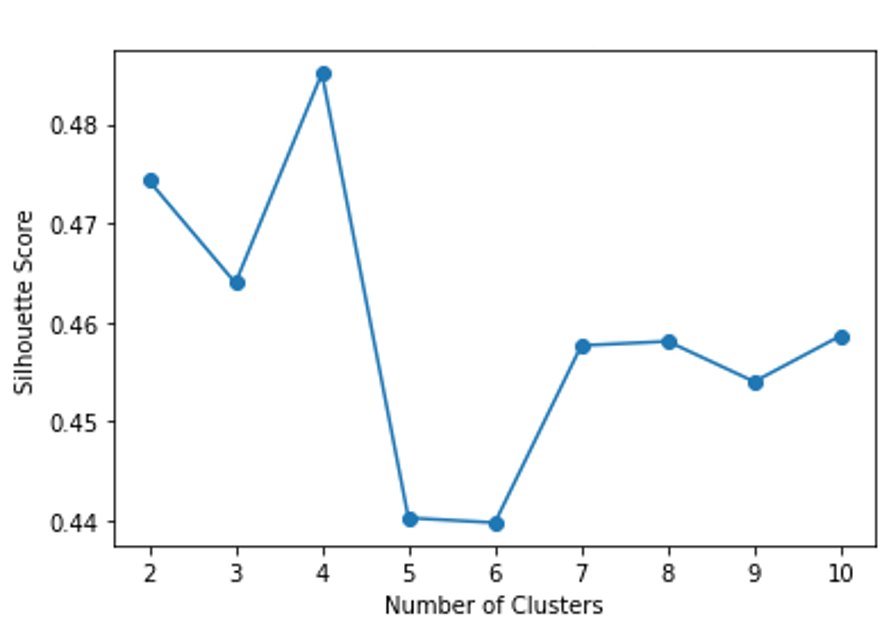}
    \caption{Silhouette Method}
    \label{fig:cluster_src_silhouette}
  \end{subfigure}
  \caption{Methods to determine the optimal value of $k$ for source domain clustering.}
  \label{fig:opt_k_src}
\end{figure}

\begin{figure}[tb]
  \centering
  \begin{subfigure}{0.48\linewidth}
    \includegraphics[width=\textwidth]{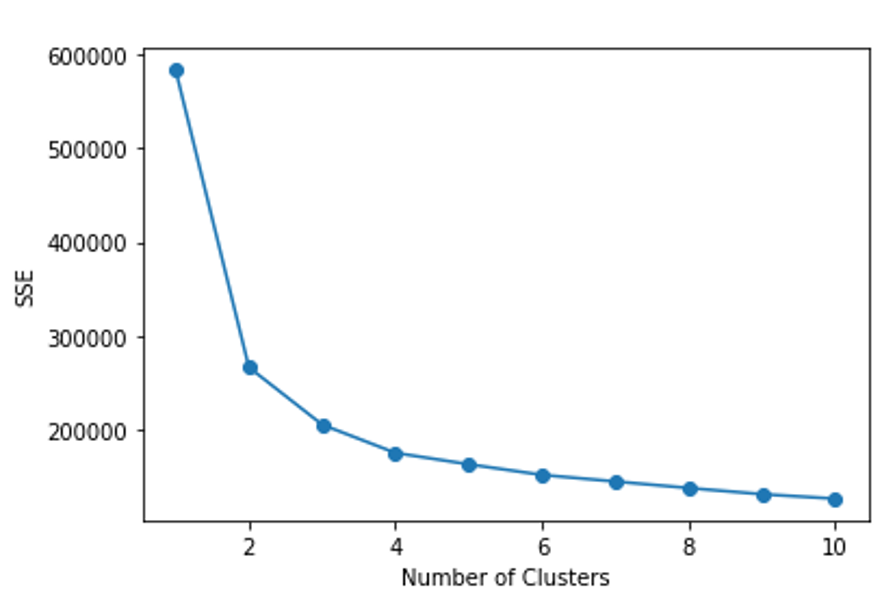}
    \caption{Elbow method}
    \label{fig:cluster_tgt_sse}
  \end{subfigure}
  \hfill
  \begin{subfigure}{0.48\linewidth}
    \includegraphics[width=\textwidth]{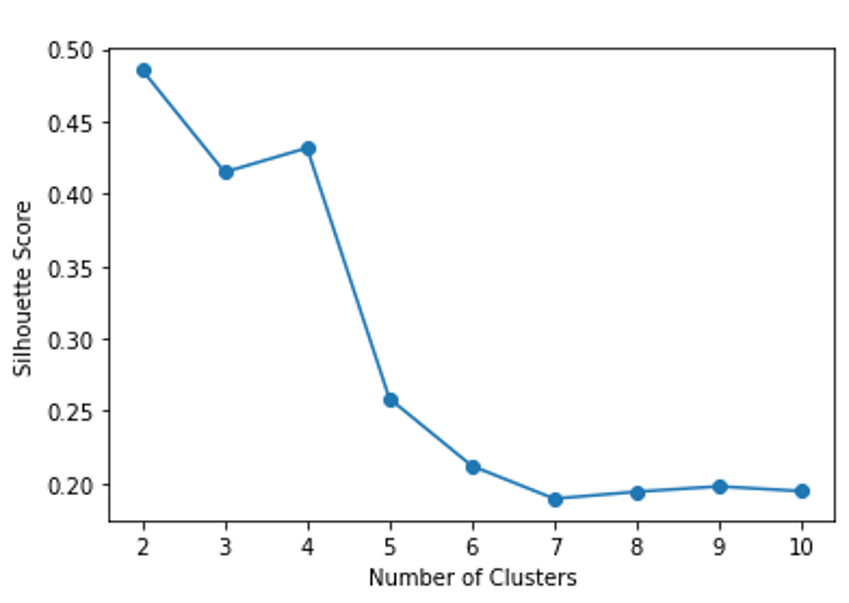}
    \caption{Silhouette method}
    \label{fig:cluster_tgt_silhouette}
  \end{subfigure}
  \caption{Methods to determine the optimal value of $k$ for target domain clustering.}
  \label{fig:opt_k_tgt}
\end{figure}

After all images from the source and target domains have been mapped to the style space, they are separately clustered. The selection of the optimal $k$ value for clustering the source and target domains is illustrated in \cref{fig:opt_k_src} and \cref{fig:opt_k_tgt}, respectively, using different methods.

\begin{figure}[tb]
  \centering
  \begin{subfigure}{0.50\linewidth}
    \includegraphics[width=\textwidth]{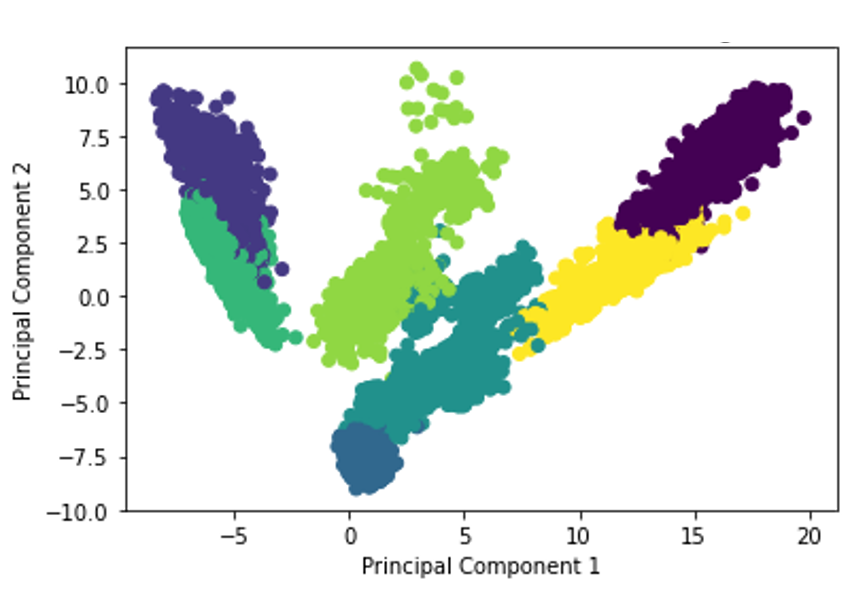}
    \caption{PCA (k=7)}
    \label{fig:cluster_src_pca_k7}
  \end{subfigure}
  \hfill
  \begin{subfigure}{0.46\linewidth}
    \includegraphics[width=\textwidth]{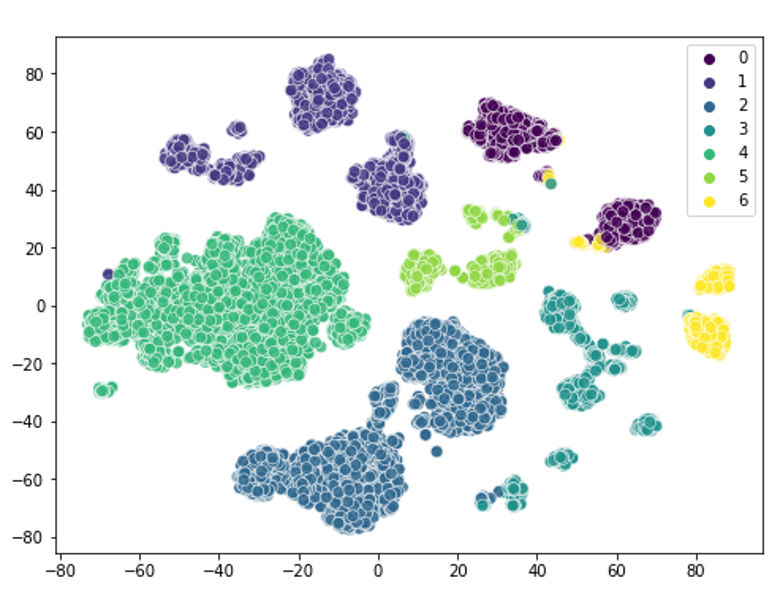}
    \caption{t-SNE (k=7)}
    \label{fig:cluster_src_tsne_k7}
  \end{subfigure}
  \caption{Visualization of source domain clustering results after dimensionality reduction.}
  \label{fig:vis_cluster_src}
\end{figure}

In \cref{fig:opt_k_src}, it is observed that according to the Elbow method, k should be chosen as 3, as it corresponds to a turning point in the curve. However, in the Silhouette method, the $k$ value with the highest Silhouette score is achieved when k equals 4. As mentioned in \cref{sec:method}, a larger $k$ value enables finer filtering of the source domain. Therefore, considering the need for more detailed clustering in the source domain, the chosen $k$ value should be no smaller than 4. After weighing the clustering time cost and Silhouette scores, the $k$ value for clustering the source domain is set to 7. The visualization of clustering results after dimensionality reduction is shown in \cref{fig:vis_cluster_src}.

\begin{figure}[tb]
  \centering
  \begin{subfigure}{0.50\linewidth}
    \includegraphics[width=\textwidth]{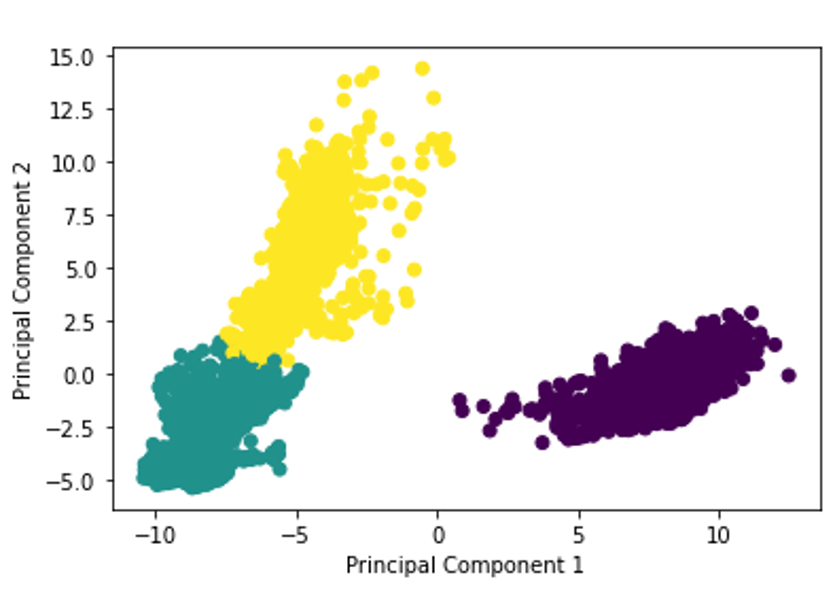}
    \caption{PCA (k=3)}
    \label{fig:cluster_tgt_pca_k3}
  \end{subfigure}
  \hfill
  \begin{subfigure}{0.46\linewidth}
    \includegraphics[width=\textwidth]{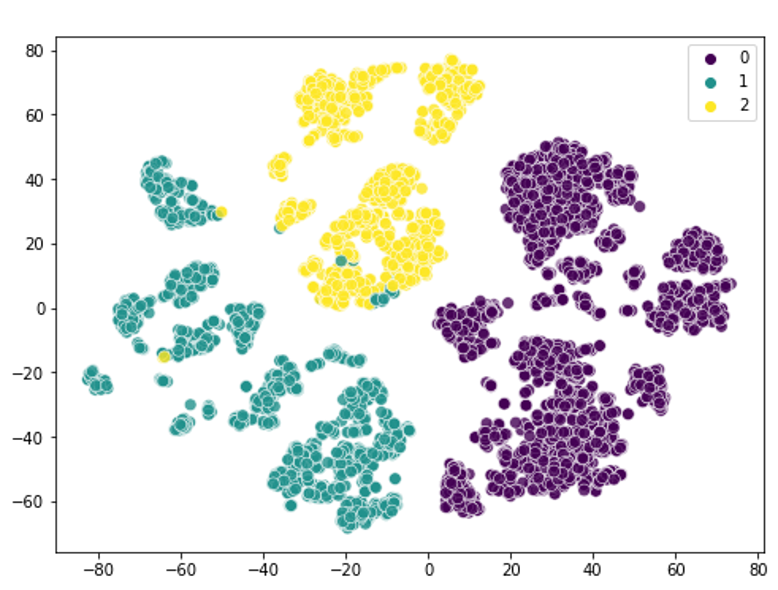}
    \caption{t-SNE (k=3)}
    \label{fig:cluster_tgt_tsne_k3}
  \end{subfigure}
  \begin{subfigure}{0.50\linewidth}
    \includegraphics[width=\textwidth]{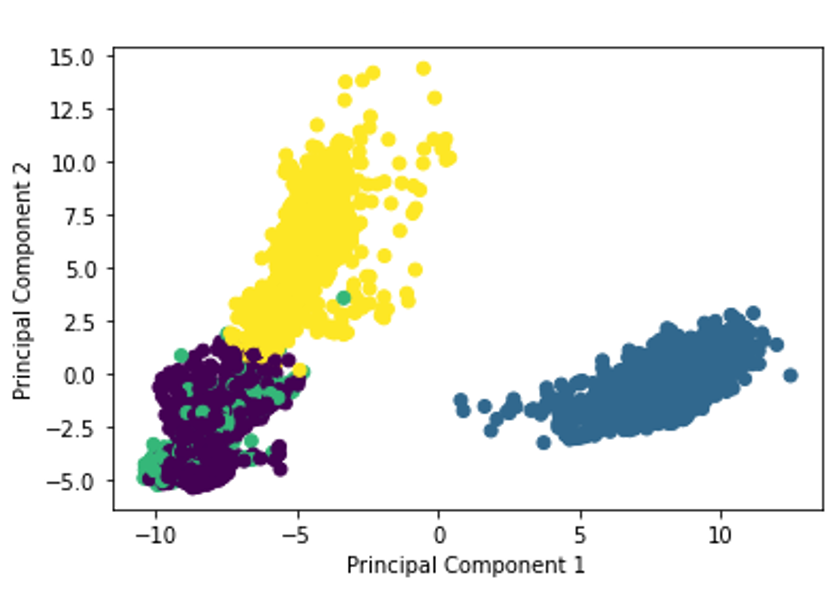}
    \caption{PCA (k=4)}
    \label{fig:cluster_tgt_pca_k4}
  \end{subfigure}
  \hfill
  \begin{subfigure}{0.46\linewidth}
    \includegraphics[width=\textwidth]{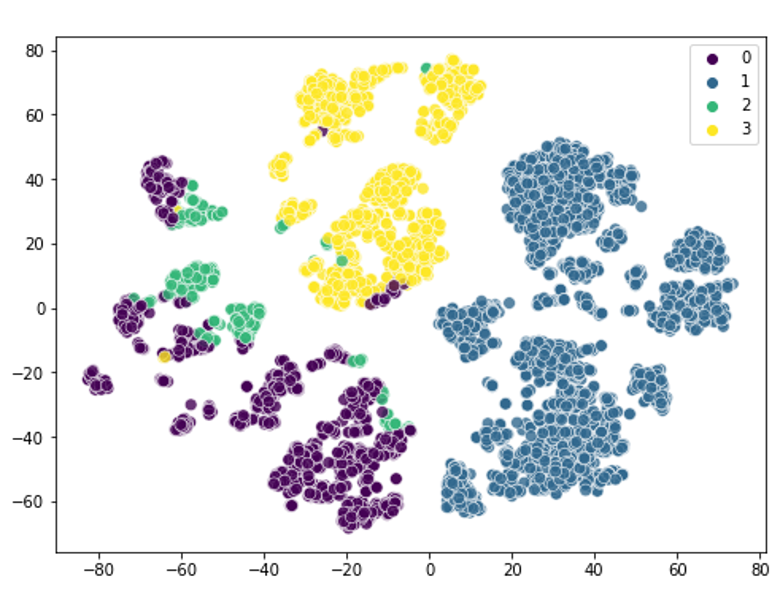}
    \caption{t-SNE (k=4)}
    \label{fig:cluster_tgt_tsne_k4}
  \end{subfigure}
  \caption{Visualization of target domain clustering results after dimensionality reduction.}
  \label{fig:vis_cluster_tgt}
\end{figure}

For the target domain, as depicted in \cref{fig:opt_k_tgt}, clustering into three or four classes appears to be a reasonable choice. Therefore, the results of clustering into three and four classes are visualized using PCA and t-SNE, as shown in \cref{fig:vis_cluster_tgt}. From the visualization, it is observed that when divided into four classes, two of the classes have a relatively high overlap. Thus, for the target domain, the $k$ value is set to 3.

\begin{figure}[tb]
  \centering
  \begin{subfigure}{0.62\linewidth}
    \includegraphics[width=\textwidth]{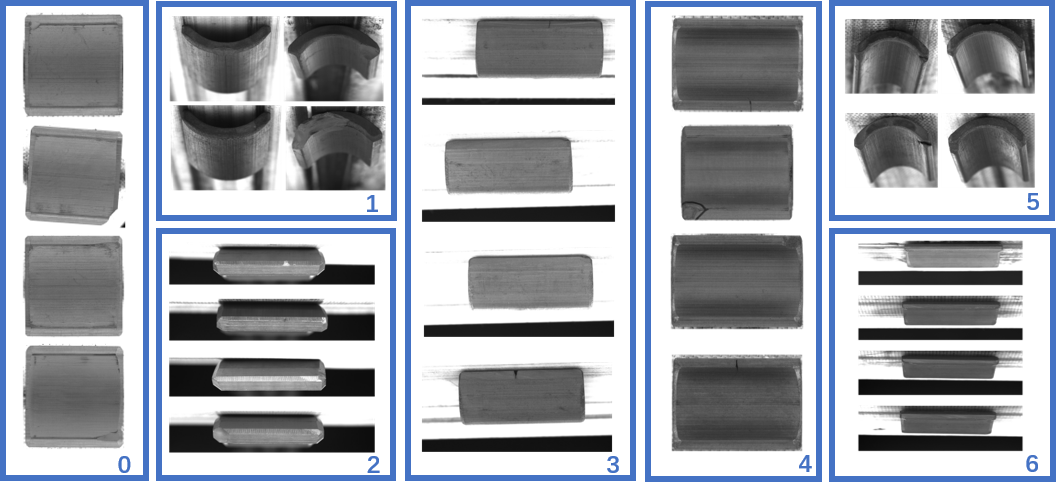}
    \caption{Source}
    \label{fig:cluster_src_instance}
  \end{subfigure}
  \hfill
  \begin{subfigure}{0.36\linewidth}
    \includegraphics[width=\textwidth]{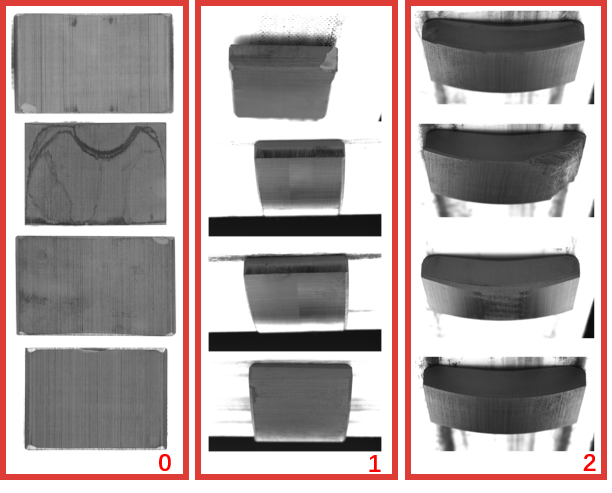}
    \caption{Target}
    \label{fig:cluster_tgt_instance}
  \end{subfigure}
  \caption{Clustering results of source and target domain sample images.}
  \label{fig:cluster_instance}
\end{figure}

\cref{fig:cluster_src_instance} and \cref{fig:cluster_tgt_instance} illustrate the mapping of clustering results back to the image space for the source and target domains, respectively. From these two figures, it is evident that the feature extraction and clustering methods employed in SF effectively classify samples from different sources with distinct differences described in \cref{fig:datasets}. The clustering results are remarkably significant. Regardless of differences in the shape design of the magnetic tiles, background, or even brightness and contrast, SF achieves excellent classification performance in both the source and target domains.



\begin{figure}[tb]
  \centering
  \begin{subfigure}{0.48\linewidth}
    \includegraphics[width=\textwidth]{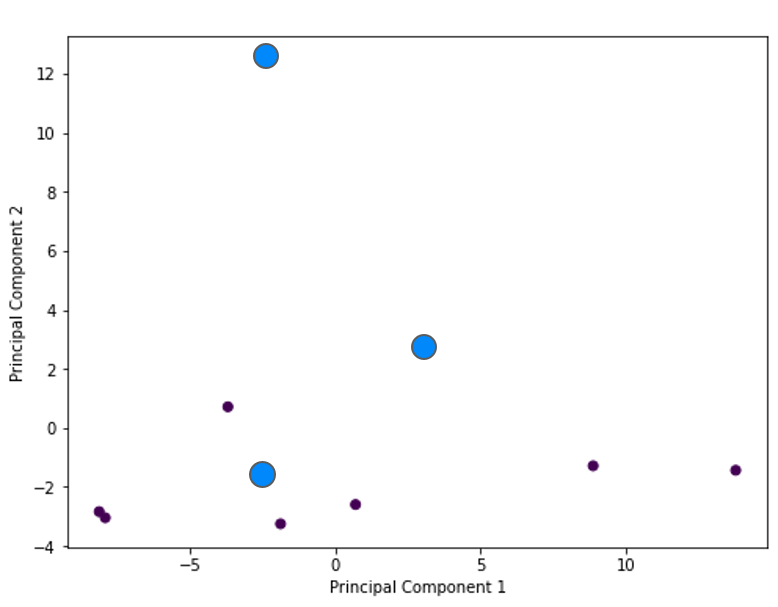}
    \caption{PCA}
    \label{fig:centroids_pca}
  \end{subfigure}
  \hfill
  \begin{subfigure}{0.48\linewidth}
    \includegraphics[width=\textwidth]{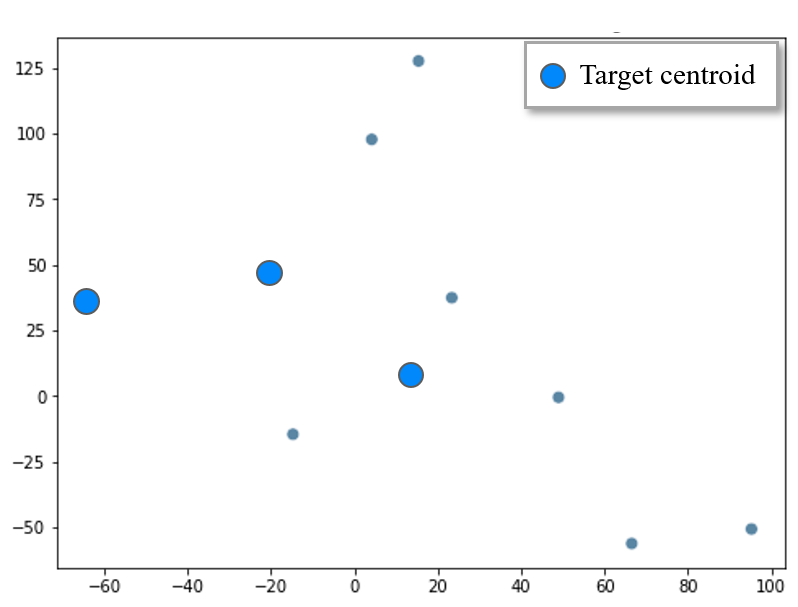}
    \caption{t-SNE}
    \label{fig:centroids_tsne}
  \end{subfigure}
  \caption{Visualization of the centroids from source and target domain in style space after dimensionality reduction.}
  \label{fig:vis_centroid}
\end{figure}

\begin{figure}[tb]
  \centering
  \begin{subfigure}{0.48\linewidth}
    \includegraphics[width=\textwidth]{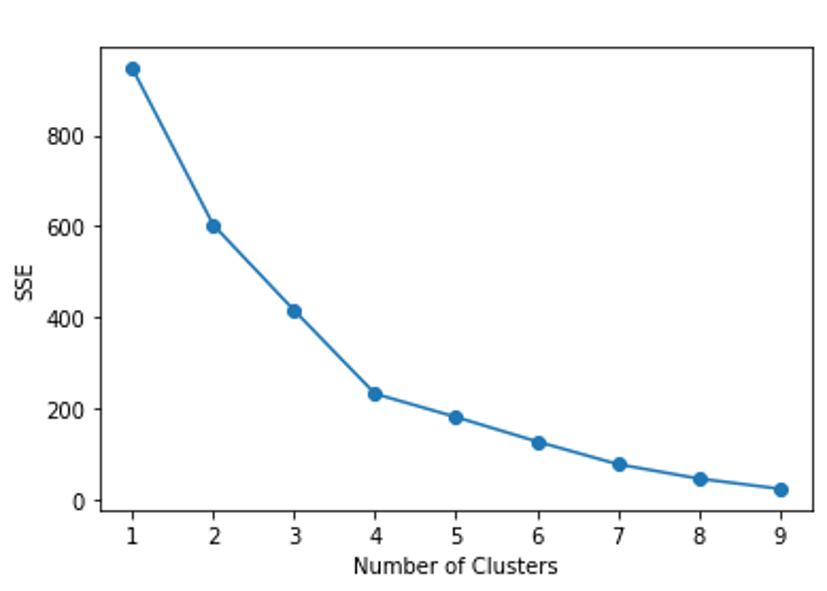}
    \caption{Elbow method}
    \label{fig:cluster_centroids_sse}
  \end{subfigure}
  \hfill
  \begin{subfigure}{0.48\linewidth}
    \includegraphics[width=\textwidth]{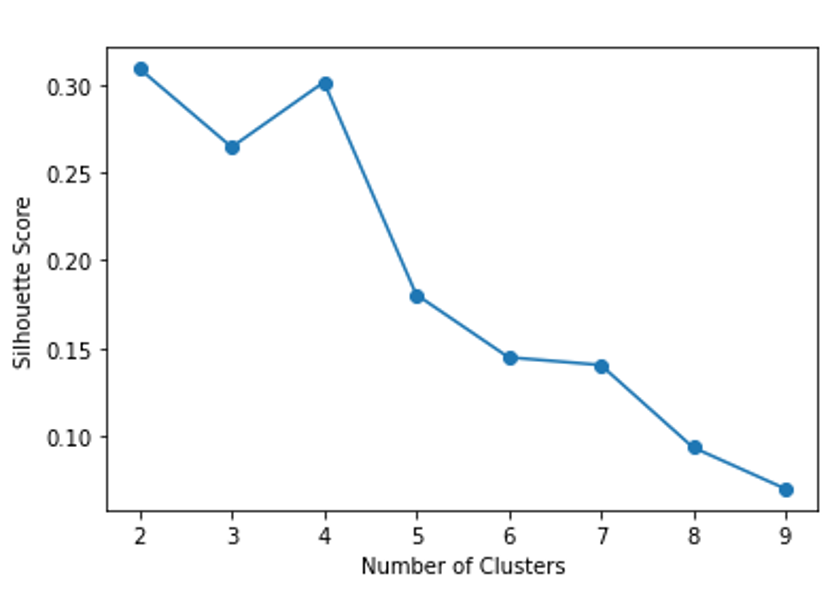}
    \caption{Silhouette method}
    \label{fig:cluster_centroids_silhouette}
  \end{subfigure}
  \caption{Find the optimal k for clustering the centroids.}
  \label{fig:opt_k_centroids}
\end{figure}

After obtaining the clustering results, the cluster centroids are computed for both the source and target domains. The dimensionality-reduced representation of the cluster centroids in the style space for both domains is depicted in \cref{fig:vis_centroid}. Subsequently, these ten centroids (seven from the source domain and three from the target domain) are further clustered. Similarly, the Elbow and Silhouette methods are utilized to select the optimal $k$, as illustrated in \cref{fig:opt_k_centroids}. Selecting $k$ as 2 or 4 would be preferable choices, as they exhibit both high Silhouette scores and correspond to the elbow points. The clustering results for centroids are presented in \cref{table:centroid_cluster}. From \cref{table:centroid_cluster}, centroids labeled 2 and 4 in the source domain are consistently not assigned to the same cluster as any centroids in the target domain. Hence, they are filtered out, resulting in the removal of the corresponding clusters along with all instances within these clusters.

\begin{table}[tb]
  \caption{Centroid clustering results. The numbers 0 to 6 represent the labels for centroids in the source domain, while the numbers 7 to 9 represent the labels for centroids in the target domain.}
  \label{table:centroid_cluster}
  \centering
  \begin{tabular}{@{}ll@{}}
    \toprule
    $k$& Results\\
    \midrule
    2 & [0, 1, 3, 5, 6, \textbf{7}, \textbf{8}, \textbf{9}], [2, 4]\\
    4 & [\textbf{7}], [0, 3, 5, \textbf{8}, \textbf{9}], [1, 6], [2, 4]\\
  \bottomrule
  \end{tabular}
\end{table}

\subsection{Validation on Transfer Learning Strategies}

Upon obtaining the source domain dataset processed by the Style Filter, we will evaluate several commonly used transfer learning strategies in industrial applications using the processed source domain dataset and a small amount of annotated target domain data. These strategies include pre-training and fine-tuning (Ptft), as well as domain adaptation (DA).

Pre-training and fine-tuning is a model-based transfer learning strategy \cite{li2022perspective} where the model first learns knowledge from the source domain extensively and then fine-tunes itself with a small amount of data from the target domain. This ensures that the model can effectively utilize the knowledge from the source domain while acquiring knowledge from the target domain, ultimately performing well on the target domain.

Domain adaptation, on the other hand, is a feature-based transfer learning strategy \cite{li2022perspective}. In \cite{li2023dayolov5}, a plug-and-play DA module is designed, which achieves domain adaptation by extracting and aligning features. In our experiment, we deploy this module onto YOLOv7 to ensure fair comparison among different transfer learning strategies, unaffected by the choice of the base network.

The validation results are presented in \cref{table:ts500} for the case where the target domain has 500 training samples, and in \cref{table:ts} for varying target domain training sample sizes from 500 to 2500 in increments of 500.

In \cref{table:ts500}, the performance comparison between Ptft and DA methods is shown when the LTS is 500, using both Style Filter-processed and unprocessed source domain datasets. From the table, it can be observed that after the SF process, the performance of Ptft significantly improves, with an increase of approximately one percentage point in both mAP@.5 and mAP@.5:.95. However, for DA, the model performance remains unchanged before and after SF. This indicates that although some source domain samples are filtered out, resulting in a reduction in the training dataset size, the model performance does not degrade; instead, it remains stable or even improves slightly. Moreover, comparing SF+Ptft with DA, we observe that the addition of SF leads to an improvement in Ptft performance, approaching that of DA and even surpassing it in the mAP@.5 metric.

\begin{table}[tb]
  \caption{Transfer learning performance comparison before and after SF.}
  \label{table:ts500}
  \centering
  \resizebox{\columnwidth}{!}{
  \begin{tabular}{@{}llllllll@{}}
    \toprule
    LTS=500 & chamfer & multifaceted & bump & impurity & crack & grind & mAP@.5/.5:.95\\
    \midrule
    Ptft & 0.800/0.516 & 0.944/0.554 & 0.933/0.704 & 0.798/0.573 & 0.721/0.464 & 0.977/0.747 & 0.862/0.593\\
    SF+Ptft & 0.770/0.498 & 0.945/0.540 & 0.937/0.722 & 0.832/0.634 & 0.768/0.496 & 0.973/0.759 & \textbf{0.871/0.608}\\
    \midrule
    DA & 0.597/0.363 & 0.927/0.515 & 0.967/0.773 & 0.862/0.674 & 0.856/0.573 & 0.987/0.778 & 0.866/0.613\\
    SF+DA & 0.611/0.370 & 0.927/0.518 & 0.960/0.769 & 0.851/0.662 & 0.858/0.573 & 0.984/0.791 & 0.865/\textbf{0.614}\\
  \bottomrule
  \end{tabular}}
\end{table}

In the experiments presented in \cref{table:ts}, we adjusted the value of LTS to observe the impact of increasing LTS on the performance of transfer learning strategies in detection tasks after the addition of the SF. From the table, it is evident that as LTS increases, the improvement in performance of the Ptft method after deploying SF is generally significant. For DA, as LTS increases, the enhancement in model performance due to SF gradually increases, with the mAP@.5:.95 metric improving by 0.1\%, 0.2\%, 0.3\%, 0.6\%, and 0.8\% as LTS increases from 500 to 2500. This indicates that for the domain adaptation strategy, SF can effectively filter out source domain data that negatively impacts the detection accuracy of the target domain, thereby enhancing the model's focus on the target domain (as LTS increases). Simultaneously, during the model training, SF reduces the attention dispersion caused by those source domain data with significantly different styles from the target domain.

\begin{table}[tb]
  \caption{Transfer learning performance comparison before and after SF with different LTS.}
  \label{table:ts}
  \centering
  \begin{tabular}{@{}llllll@{}}
    \toprule
    LTS & 500 & 1000 & 1500 & 2000 & 2500\\
    \midrule
    Ptft & 0.862/0.593 & 0.917/0.659 & 0.947/0.732 & 0.958/0.759 & 0.967/0.785\\
    SF+Ptft & \textbf{0.871/0.608} & \textbf{0.923/0.666} & \textbf{0.948/0.740} & 0.956/\textbf{0.759} & \textbf{0.969/0.792}\\
    \midrule
    DA & 0.866/0.613 & 0.896/0.655 & 0.908/0.672 & 0.914/0.683 & 0.925/0.695\\
    SF+DA & 0.864/\textbf{0.614} & \textbf{0.901/0.657} & \textbf{0.910/0.675} & \textbf{0.922/0.689} & \textbf{0.931/0.703}\\
  \bottomrule
  \end{tabular}
\end{table}

\section{Discussion} 



Through experimental validation, we observed a phenomenon: SF remains stable improvement in Ptft, but becomes more pronounced in DA as LTS increases. Fundamentally, our source and target domain data differ only in terms of factors such as brightness, contrast, texture and background due to different shooting environments, while they share the same target types (defects conform to industry standards). The role of our Style Filter is to filter and remove samples from the source domain that differ significantly from the target domain, a difference that can be described by Style, as confirmed with an experiment in \cref{fig:style}. The reason why filtering and removing samples from the source domain is effective lies in the fact that within these differences describable by style, there exist similarities or overlaps with certain aspects of the defect features that need to be detected and extracted in the detection task. This implies that samples with excessive style differences will affect the network's focus on and learning of defect features, leading to attention dispersion and negatively impacting the model's performance in achieving good results in the target domain task through learning. Therefore, the role of SF can be described as reducing interference with the network's learning task objectives (defects) by filtering data.

From another perspective, SF can aid the model in directing its attention more towards defects. In the case of Ptft, during the pre-training, the portion of source domain data removed by SF can lead the network to reduce its focus on overcoming various styles of samples and instead pay more attention to extracting and discriminating the features of defects. Regarding DA, a dual-stream network, the model strives to maintain robustness to styles while learning task-relevant features (defects). However, the removal of source domain data by SF may cause the network to overly focus on styles rather than defects, affecting its ability to detect defects.

\section{Conclusion}

In this work, a tailored methodology named Style Filter is introduced for industrial contexts. By selectively filtering source domain data before knowledge transfer, Style Filter reduces the quantity of data while maintaining or even enhancing transfer learning performance. Offering label-free operation, minimal reliance on prior knowledge, independence from specific models, and re-utilization, Style Filter is evaluated on authentic industrial datasets, highlighting its effectiveness when employed prior to conventional transfer strategies in the deep learning domain. The results underscore the efficacy of Style Filter in real-world industrial applications.


%
%
\bibliographystyle{splncs04}
\bibliography{main}
\end{document}